\documentclass[nohyperref]{article}

\usepackage{microtype}
\usepackage{graphicx}
\usepackage{booktabs} 
\usepackage{listings} 
\graphicspath{ {./images/} } 

\usepackage{hyperref}

\usepackage[accepted]{icml2022}

\usepackage{amsmath}
\usepackage{amssymb}
\usepackage{mathtools}
\usepackage{amsthm}
\usepackage{listings}
\usepackage{color}
\usepackage{xcolor}
\usepackage{float}
\usepackage{caption}
\usepackage{subcaption}

\definecolor{eclipseStrings}{RGB}{42,0.0,255}
\definecolor{eclipseKeywords}{RGB}{127,0,85}
\colorlet{numb}{magenta!60!black}

\lstdefinelanguage{json}{
    basicstyle=\normalfont\fontsize{8.5}{9.5}\selectfont\ttfamily,
    commentstyle=\color{eclipseStrings}, 
    stringstyle=\color{eclipseKeywords}, 
    numbers=left,
    numberstyle=\scriptsize,
    stepnumber=1,
    numbersep=0pt,
    showstringspaces=false,
    breaklines=true,
    frame=lines,
    string=[s]{"}{"},
    comment=[l]{:\ "},
    morecomment=[l]{:"},
    captionpos=b,
    breaklines=true,
    literate=
        *{0}{{{\color{numb}0}}}{1}
         {1}{{{\color{numb}1}}}{1}
         {2}{{{\color{numb}2}}}{1}
         {3}{{{\color{numb}3}}}{1}
         {4}{{{\color{numb}4}}}{1}
         {5}{{{\color{numb}5}}}{1}
         {6}{{{\color{numb}6}}}{1}
         {7}{{{\color{numb}7}}}{1}
         {8}{{{\color{numb}8}}}{1}
         {9}{{{\color{numb}9}}}{1}
}

\definecolor{deepblue}{rgb}{0,0,0.5}
\definecolor{deepred}{rgb}{0.6,0,0}
\definecolor{deepgreen}{rgb}{0,0.5,0}
\DeclareFixedFont{\ttb}{T1}{txtt}{bx}{n}{8.5} 
\DeclareFixedFont{\ttm}{T1}{txtt}{m}{n}{8.5}  
\newcommand\pythonstyle{\lstset{
language=Python,
basicstyle=\ttm,
morekeywords={self},              
keywordstyle=\ttb\color{deepblue},
emph={MyClass,__init__},          
emphstyle=\ttb\color{deepred},    
stringstyle=\color{deepgreen},
frame=tb,                         
showstringspaces=false,
numbers=left,
numberstyle=\scriptsize,
numbersep=3pt,
breaklines=true,
captionpos=b,
commentstyle=\color{gray}, 
}}

\lstnewenvironment{python}[1][]
{
\pythonstyle
\lstset{#1}
}
{}

\usepackage[capitalize,noabbrev]{cleveref}

\theoremstyle{plain}

\theoremstyle{definition}

\theoremstyle{remark}

\usepackage[T1]{fontenc}

\usepackage[textsize=tiny]{todonotes}

\icmltitlerunning{Submission and Formatting Instructions for ICML 2022}

\begin{document}

\twocolumn[
\icmltitle{feather - a Python SDK to share and deploy models}



\icmlsetsymbol{equal}{*}

\begin{icmlauthorlist}
\icmlauthor{Nihir Vedd}{equal,icl}
\icmlauthor{Paul Riga}{equal,ef}
\end{icmlauthorlist}

\icmlaffiliation{icl}{Department of Computing, Imperial College London, United Kingdom}
\icmlaffiliation{ef}{Undisclosed}

\icmlcorrespondingauthor{Nihir Vedd}{nihirvedd@gmail.com}
\icmlcorrespondingauthor{Paul Riga}{riga\_paul@hotmail.com}

\icmlkeywords{Machine Learning, ICML}

\vskip 0.3in
]
\thispagestyle{plain}
\pagestyle{plain}


\printAffiliationsAndNotice{\icmlEqualContribution} 

\begin{abstract}
At its core, feather was a tool that allowed model developers to build shareable user interfaces for their models in under 20 lines of code. Using the Python SDK, developers specified visual components that users would interact with. (e.g. a FileUpload component to allow users to upload a file). Our service then provided 1) a URL that allowed others to access and use the model visually via a user interface; 2) an API endpoint to allow programmatic requests to a model.

In this paper, we discuss feather's motivations and the value we intended to offer AI researchers and developers. For example, the SDK can support multi-step models and can be extended to run automatic evaluation against held out datasets. We additionally provide comprehensive technical and implementation details.

N.B. feather is presently a dormant project. We have open sourced our code for research purposes: \url{https://github.com/feather-ai/}
\end{abstract}

\section{Introduction}
\label{sec:introduction}
Over the last decade or so, research and interest in machine learning (ML) has grown exponentially. This growth has led to the development and adoption of specialized tools which provide a significantly better development experience. Typically, these tools target one part of the multi-part modelling pipeline. For example, PyTorch \cite{NEURIPS2019_9015} targets the model building itself; Weights \& Biases \cite{wandb} targets experiment tracking and model versioning.

Until 2021, very few popular tools existed that allowed model developers (MDs) to easily share their models with other people (model consumer; MC). The most common way of doing so was to upload code to a public repository (e.g. GitHub) and provide instructions to technical MCs in order to recreate a published model. Another alternative is for the MD to create a front-end that enables both technical and non-technical MCs to interact with the model. Unfortunately, in either case, the incentives for MDs are not clear. From the perspective of the MD, the former case is considerably less work. However, there are multiple drawbacks. It: 1) makes the model inaccessible to those without coding ability; 2) places a responsibility on the MD to maintain the repository; 3) requires an MC to have the relevant hardware to use the model. 

Additionally, anecdotal evidence has shown that for a technical MC to qualitatively use and test the model, hours of setup may be required. These problems can largely be solved with a custom front-end, but: 1) it requires the MD (or their team) to understand front-end and back-end technologies; 2) requires maintainers to ensure the quality of service (including scaling infrastructure when necessary); 3) is out-of-scope based on current conventions for ML research; 4) requires a financial commitment from the MD or their affiliated lab for hosting models.

\begin{figure*}[t!]
    \centering
        \includegraphics[width=0.36\textwidth]{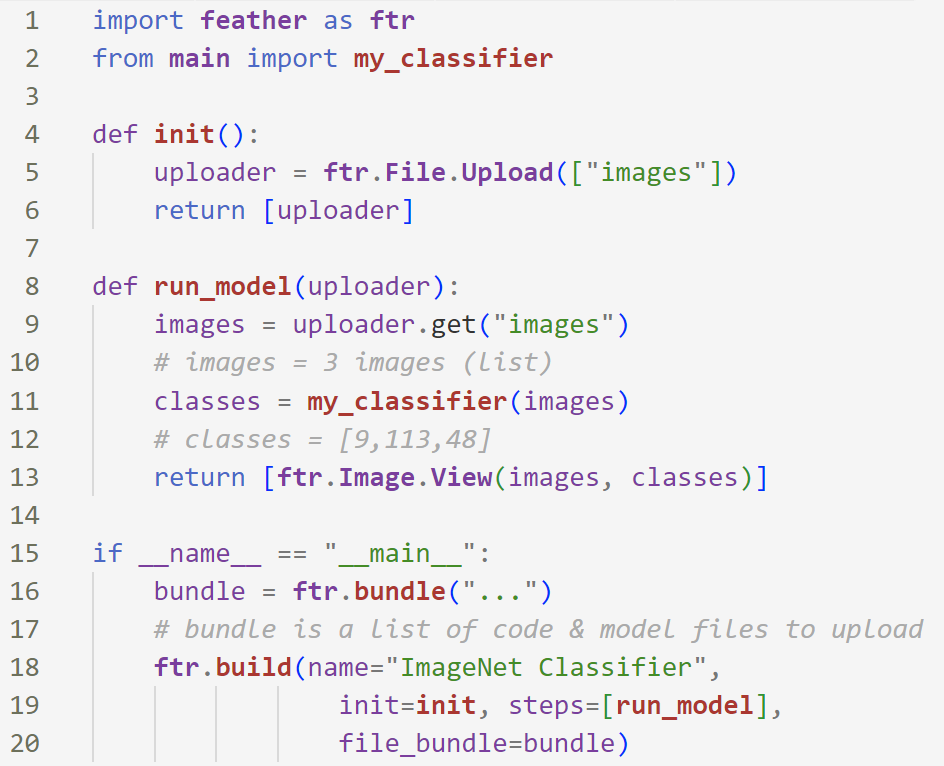}
        \includegraphics[width=0.25\textwidth]{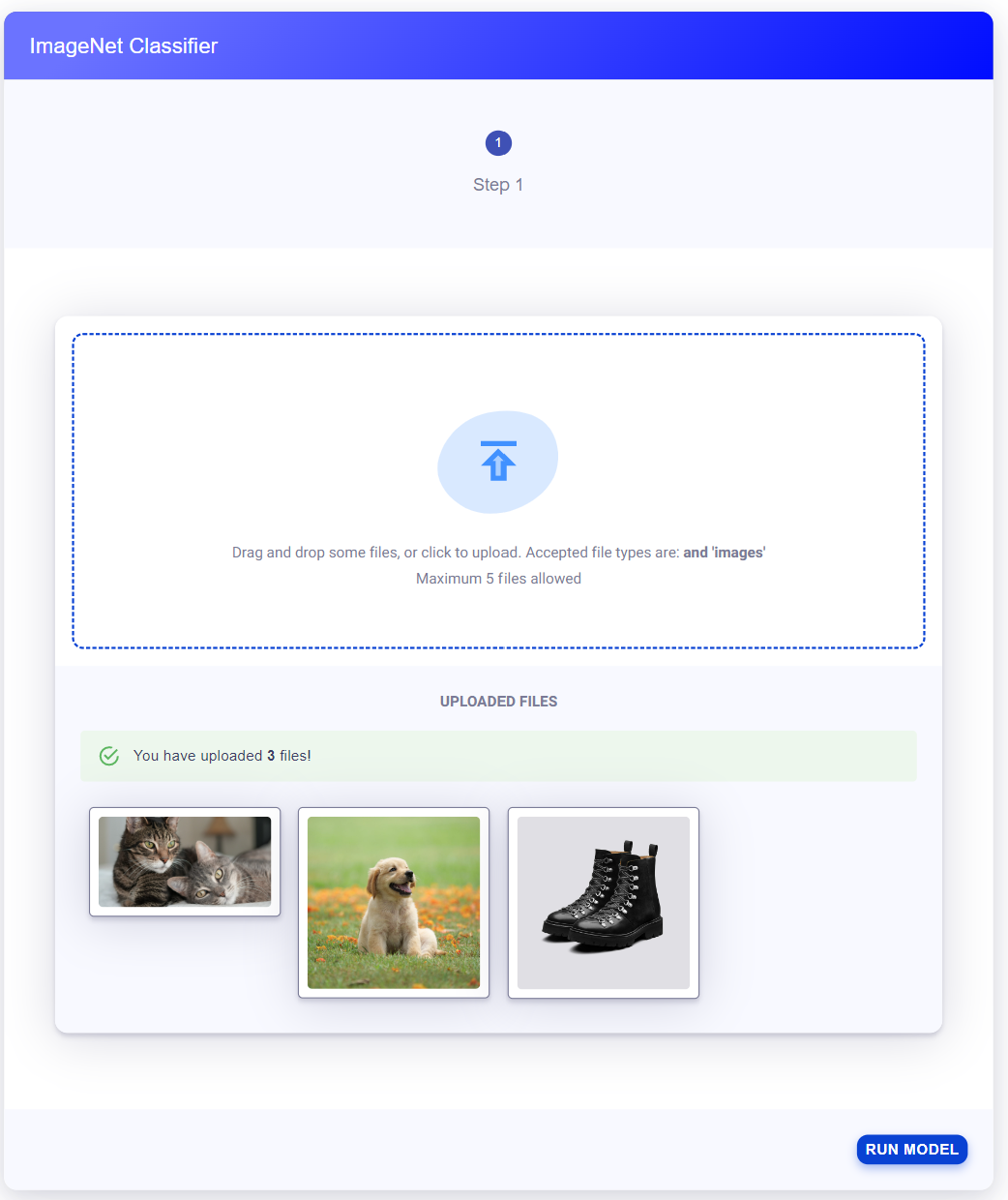}
        \includegraphics[width=0.28\textwidth]{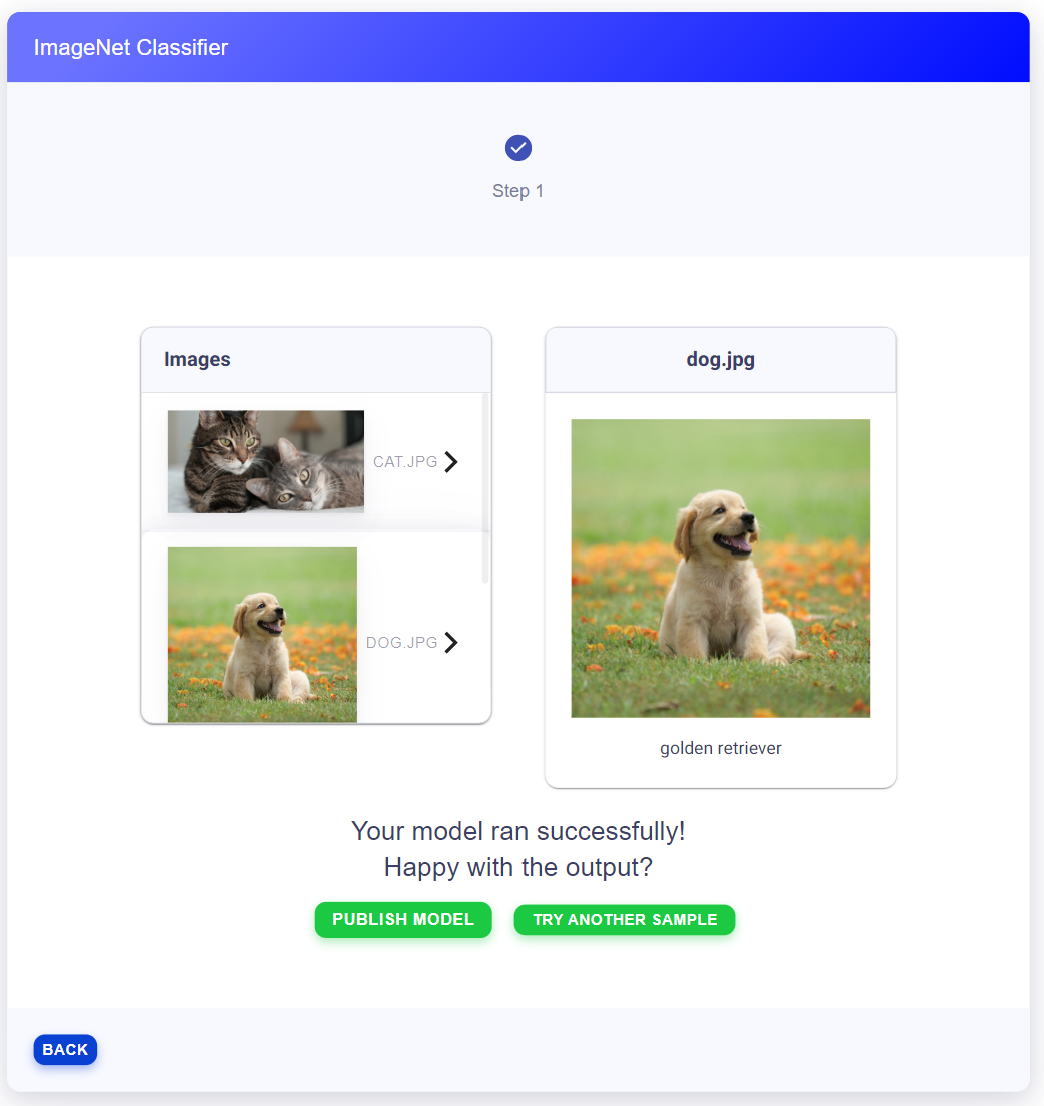}
    \caption{Left: An example feather script that deploys an image classifier. Right: The UI rendered by the feather script. `my\_classifier' (line 11) is an inference function that runs a classification model. It is regular Python code that can call any library (e.g. PyTorch, Tensorflow etc.)}
    \label{fig:script-and-ui}
\end{figure*}

The overarching effect of the above points is two-fold: 1) AI research in general loses transparency as it becomes (unnecessarily) harder to verify results; 2) MDs who are ideating may be deterred from diving deeper into their ideas as practical or technical barriers prevent them from using a model for small-sample inference, let alone training a model from scratch. Both of these points ultimately lead to a slower rate of production and adoption of AI.

At the time of feather's inception, we were able to find only 2 tools which allowed MDs to visually share their code via a component-based SDK: Gradio \cite{Abid2019Gradio:Wild} and Streamlit \cite{streamlit}. We discuss these further in Section \ref{sec:related-work}. Both tools faced restrictive drawbacks. A common drawback was the lack of support for multi-step models. That is, a model which is composed of multiple sub-models within it, where each sub-model requires human interaction. For example, a guided image captioning system (e.g. \citet{Ng2020UnderstandingDomainsb} may require a human to explicitly select objects they want captions for before generating a caption. That is, after inputting an image, an object detector outputs a list of detected objects for a human to control the caption generation upon. A human then selects a subset of objects as input to a conditional caption generation model. Otherwise, Gradio used to only allow MDs to share their models for 6 hours at a time. These models were hosted directly on the MDs machine - thus exposing a myriad of security issues. Long lived/Gradio hosted models required a premium subscription levied on the MD. HuggingFace Spaces \cite{wolf-etal-2020-transformers} was released during feather's development process. These largely solved the aforementioned issues.

Despite this, feather was envisioned to be more than a place to simply share your models via a UI. Within the direct scope of our SDK, we can:
\begin{itemize}
    \setlength\itemsep{0em}
    \item Run models through a UI
    \item Execute these models programmatically via an API. feather can also automatically generate documentation for these APIs.
    \item Deploy multi-step models (with built in state management that allows users to re-run or refresh steps)
    \item Automatically model versioning
\end{itemize}\label{features}

feather was designed to have natural extensions that could ease the deployment workflow for an MD and further democratize AI research. Specifically, we developed our SDK such that the following features could be supported. + indicates a research-based feature (we refer to these often through the paper):
\begin{itemize}
    \item Automatic evaluation against held out datasets +
    \item Provide a ModelOps like experience, where the MD can visualise incoming data to counter-act model/concept drift.
    \item Fine-tuning as a service. That is, allow MCs to upload data to fine-tune a model that an MD has published.
    \item For UI based execution, enable MCs to provide feedback on the model predictions per sample uploaded. MDs may then be able to use this now-labelled data to augment their dataset or model. Future extensions could have first-party support for RLHF \cite{Christiano2017DeepPreferences}.
    \item Optionally require MCs to pay per execution (thus financially incentivizing MDs to maintain and upkeep their models).
    \item Group models by task and/or dataset to allow easy access. Coupled with the aforementioned automatic evaluation point, this allows us to build leaderboards for different tasks and/or datasets. +
    \item Embedding models on different websites
    \item Permission sharing of models (e.g. Private/Unlisted/Public) and team-oriented features
\end{itemize}

The rest of this paper is structured as follows. In Section \ref{sec:related-work} we provide an overview of the alternate approaches to solve the aforementioned problems. In Section \ref{sec:methodology} we dive into technical and implementation details on our system. Section \ref{sec:improvements} discusses improvements directly related to the SDK.

\section{Related Work}\label{sec:related-work}
As hinted at in Section \ref{sec:introduction}, there are solutions of varying level of difficulty which allow an MD to share a model. Here, we introduce these solutions and discuss key differences where possible. Between 2021 and early-2022, when feather was actively being developed, we found only 3 tools that had considerable community adoption: Gradio, Streamlit, and Replicate. 

\subsection{Gradio}\label{sec:gradio}
Gradio's \cite{Abid2019Gradio:Wild} core contribution is similar to feather's - a simple-to-use SDK that allows researchers to build shareable interfaces/webpages for their models. At the time of feather's development, Gradio lacked support for features that feather provided out of the box: Server-based hosting (N.B. this was a paid feature); multi-step execution; API-based execution. In order, server-based execution allows an MD to upload their models to an external server as opposed to hosting the model on their own machine. This solves an array of security flaws and enables a model to be accessed at any time, instead of the temporary 6 hours connection provided by Gradio. Examples of security flaws may involve a malicious MD storing personal information uploaded by an MC. They may also involve a malicious MC entering eval statements in the MDs code; since the models are not executed in a sandboxed environment, this may pose a threat to the MDs machine. However, HuggingFace Spaces now solves the problem of self-hosting. (discussed further in Section \ref{sec:hface_spaces}). Multi-step execution enables deploying models that require at least 1 human interaction in the model pipeline. Examples may include \citet{Vedd2021GuidingGeneration} and \citet{Jiang2022Text2Human:Generation}. API-based execution allows a model to be used programmatically for high-throughput inference.

\subsection{Streamlit}
Streamlit \cite{streamlit} is a product for sharing data apps. Similar to Gradio and feather, Streamlit allows you to define components in Python code to be rendered onto a webpage. Streamlit focuses its effort on  single-step execution. At some point during feather's development, Streamlit released its ``state'' feature. The state feature enables MDs to run callback functions upon an input change. Thus, it may be utilized execute models which require multiple steps.

Streamlit's value as a place to build web interfaces and share models is clear. However, it lacks the support for programmatic execution of models via an API. As a result, you cannot automatically evaluate models against a dataset. Additionally, it positions itself as a product and has shows no intention to support the research based features outlined in the features in Section \ref{features}.

\subsection{HuggingFace Spaces} \label{sec:hface_spaces}
Around mid 2021, HuggingFace Spaces \cite{wolf-etal-2020-transformers} (henceforth Spaces) was released. Spaces allows hosting and sharing of ML apps made by community members or organisations. Crucially, it itself is not an SDK used to create interfaces for models. Rather, it is a service that allows MDs to upload their models which have interfaces created by Streamlit or Gradio. Spaces alleviates the self-hosting deployment procedure of Gradio (see Section \ref{sec:gradio}) and provides a user with a clean URL to their model (e.g. huggingface.co/spaces/<username>/<modelname>). With respect to UI-based execution, feather = Spaces + Gradio.

Spaces allows execution on CPU or GPU. CPU execution is always free, however GPU execution levies a cost on the MD. One difference between feather and Spaces is that we attempt to reverse this cost by levying a cost on the MC. Another difference is that Spaces only allows UI-based execution of models. Similar to the drawbacks of Streamlit, this means that many of the research based features outlined in Section \ref{features} are not supported by Spaces.

\subsection{Replicate}
Replicate (organisation) \cite{replicate} consists of 2 different sub-products: Cog and the web-based tool Replicate. Cog is an open-source tool that allows MDs to containerize ML models. Cog attempts to simplify the deployment process of an ML model through multiple means. Alongside the containerisation process, an MD can define the inputs and outputs of their model in Python, from which Cog automatically creating an OpenAPI schema. Cog will then use this to automatically create a RESTful API. Cog additionally provides some deployment features, such as queue workers and cloud storage. Though their roadmap does not indicate support for the research based features we outlined in Section \ref{features}, their API-first philosophy means their product may be extended to support these. Finally, Cog does not support multi-step models.

Recently, Replicate (organisation) released their web-based product, Replicate. Replicate enables UI-based execution of a Cog container. We omit further discussion about Replicate as during feather's development and release, the web-based Replicate did not exist.

\subsection{Self deployment}
The most flexible approach of deploying a model is self-deployment and hosting. This allows an MD to create interfaces which are fully bespoke to the model at hand. Though this category of deployment does both allow API and UI-based execution of models, there are drawbacks to this approach from both a product perspective and an MDs perspective. From a product perspective, it is non-trivial to compare the model the MD has created to other models across different websites. Hence research based features such as leaderboards and discoverability of models is severely limited.

From the MD perspective, there is a significant trade-off in flexibility vs development time. Ignoring any financial costs, the time cost of building an interface from scratch is lengthy, and many use different technologies than what an MD is familiar with already. Management and scalability of models and infrastructure is now also the MDs responsibility. Additionally, many features which feather planned to support (e.g. data visualisation for the MD; model feedback from the MC to MD) are features that would need to explicitly be built by the MD.

From an MCs perspective, they now have to discover the different websites that are create by MDs - a central place to find models would not exist. Additional friction is also placed on the MC to register and enter payment details for each different custom website/product/service they use.

\section{User Research}
feather's feature set was developed after conducting a dozen 1 hour interviews with ML practitioners. Strictly, practitioners are MDs, however our research uncovered that many MDs may also be MCs. The practitioners ranged a variety of professions. Some were researchers, some hobbyists, and others worked in industry settings. They had varying use-cases for a deployment tool such as feather, however one common pain point that was highlighted was that of accessing and using models that have been shared by others. Solving this problem was a core motivation for feather.

The rest of this section compiles the findings we made during our research process. We structure our learnings as follows. We find segmenting the practitioners by profession is most natural. For each segment, we present general learnings for the following questions: 1) Under what context do you share your models; 2) How do you share them; 3) How do you maintain them; 4) Does your work entail you using models that others have developed; 5) What problems do you face when using other peoples models.

\textbf{Researchers} typically share their models alongside paper publications, despite this not being a strict requirement for accepted work. They are motivated by encouraging innovation in the scientific community, and value open source access because this is normally something that they have benefited from as well. Their models are normally publicly shared on a version control website such as GitHub, where source files and training instructions are provided. Researchers may or may not release model weights. They often have intentions to maintain their code, but find that unless someone opens an issue, the code will stay as-is. Researchers often utilise models developed by other researchers to either extend the previous researchers contributions, or to test the models capabilities. In other words, they are both MDs and MCs. Typically, researchers are frustrated at the disproportionate time it takes to run inference on models that other researchers have released. This `disproportionate' time exists for 2 reasons: 1) Issues with environments; 2) Lack of README and/or documentation in the released code.

\textbf{Hobbyists} are individuals who explore and utilise machine learning models out of personal interest or for smaller-scale projects. They often share their models by building a custom interface and hosting the binaries on a cloud-based service such as AWS. MCs may find these models through the blogs, forums, or social media platforms that the hobbyists share them on. Hobbyists also leverage existing open-source models and libraries and contribute to them when possible. The motivation of up-keeping a model for a hobbyists varies on a person-to-person case. However, the following reasons seem to be emerge as possible motivations: financial returns; being part of a technical community; personal learning. Hobbyists frequently use models developed by others, as they often seek inspiration and learn from existing projects. They may modify or extend these models to suit their own use-cases or simply experiment with them to gain a deeper understanding. Similar to researchers, they often experience problems setting up environments or lack of documentation. Additionally, hobbyists may not have the same levels of resources as researchers or industry professionals. Thus, executing larger models often proves challenging.

\textbf{Industry Professionals} work in various sectors, including healthcare, finance, and technology, and they utilise machine learning models to solve real-world problems. These professionals typically share their models internally within their organization or with clients. Whilst in the development process, the most common way MDs `share' their model with other internal stakeholders is screen sharing and interacting with the model via CLI. If multiple people are working on the model code, version control software such as GitHub is utilised. Once developed, MDs in industry use either proprietary platforms or cloud-based services for model sharing and deployment. Industry professionals have a strong focus on maintaining and updating their models to ensure optimal performance and to meet business requirements. Some businesses also invest in ModelOps, ensuring that consumer facing models remain at an adequate quality. At the time of conducting these interviews, they often used models developed by others, either from research, other companies, or even internally. Some professionals in this category did write custom model code themselves. Where possible, maintained services were preferred because these types of models have been empirically validated by other clients. Their largest concern with other peoples models were the lack of robustness, and potential real world failure cases which have not been identified in the associated research paper.

As seen above, the problems these three segments of practitioners face when using other people's models are diverse, yet some common themes emerge. Environment setup issues and lack of documentation are prevalent across all three groups, indicating that more streamlined methods for sharing models and clearer instructions on usage could significantly improve the user experience. Additionally, concerns about robustness, resource requirements, and real-world applicability of shared models were noted.

By understanding these common pain points, feather's feature set was designed to address the specific needs of each user segment while also tackling the broader issues that affect the entire machine learning community. feather was thus motivated to be a tool that simplified model sharing, deployment, and maintenance.

\section{Methodology}\label{sec:methodology}
\begin{figure}[!ht]
    \includegraphics[width=\linewidth]{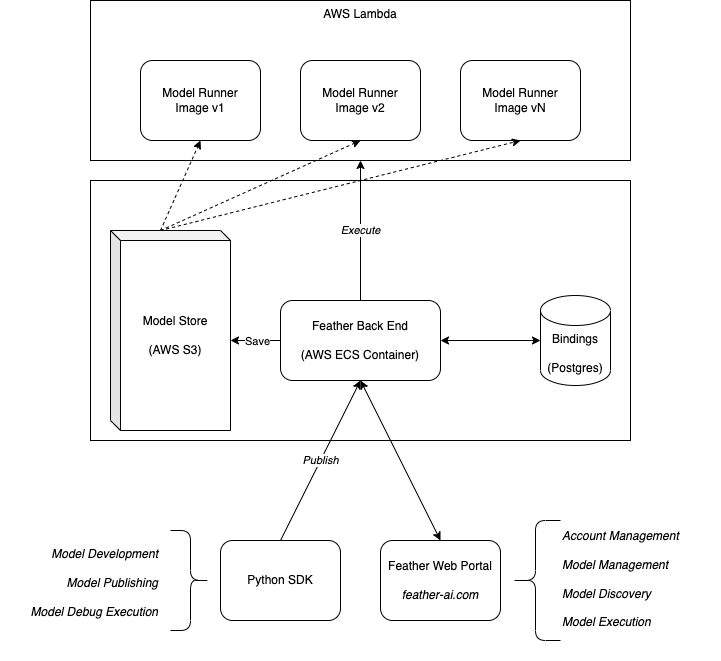}
    \caption{The feather architecture is centred around 2 key functions: model creation/publishing and model execution. The high level system diagram is shown here and comprises the following key components: \\ The \textbf{Python SDK} is a client-facing python library. It implements feather’s component APIs which allows MCs to quickly create multi-step models. Rich metadata about the model is generated, and this information is later used when executing the model. \\ The \textbf{Web Portal} is single page application which allows unauthenticated users to browse and execute available models. Authenticated users can use the portal to manage their published models. The execution requires utilising the per-model metadata (created by the SDK during publishing) to generate the UI dynamically for each model. \\ \textbf{feather's back end} associated database, provides server-side functionality via a REST API. Both the SDK and the Web Portal talk directly with this service (for authentication, data validation, etc) to perform all actions on the feather platform. \\ The \textbf{model runner} is an AWS Lambda image that is dynamically created from a Base Image for each model + dependencies type. The runner is loaded as needed to execute an MDs model. During execution, the image is loaded and executed by the Lambda runtime in a sandbox environment for safe execution of user-provided code.}
    \label{fig:feather-archictecture}
\end{figure}

In this section we discuss underlying implementation details of feather. To be explicit, feather is library agnostic. It is a system that publishes and executes code. Any framework (e.g. PyTorch, Tensorflow, SKlearn etc.) may be utilised in the publisher's code. In other words, feather’s architecture facilitates 2 key functions: model creation/publishing and model execution. To achieve this, our system requires 4 components: an SDK (Section \ref{sec:sdk}), a front-end web application and a model runner (Section \ref{sec:running}), and a back-end service (Section \ref{sec:service}) which ties the system components together. A diagram of how the different components interact with each other can be seen in Figure \ref{fig:feather-archictecture}.

Reader's are likely aware that ML models are based on mapping an input to an output. For example, an input of a handwritten digit can be mapped to a class label of the digit, or an input of a random noise vector can be mapped to some high-dimensional image. Some models may also require multiple steps of execution before generating a final output. For example, \citet{Jiang2022Text2Human:Generation}, \citet{Ng2020UnderstandingDomainsb} and \citet{Vedd2021GuidingGeneration}. In the latter, the authors devise a visual question generation (VQG) algorithm that has three steps: 1) Input an image; 2) An object detector outputs a set of detected objects in the image; 3) A question generator uses a selected subset of these detected objects to generate a natural language question. Aside from self deployment, the tools we outlined in Section \ref{sec:related-work} do not support such a multi-step use case. Hence, we motivate the sub-sections below with this example.

\subsection{SDK}\label{sec:sdk}
This component is a client-facing Python library that is publicly available. The SDK implements feather’s component APIs which allows MDs to quickly share/deploy (multi-step) models. The SDK is a core component of feather, as it is responsible for gathering rich metadata about the model, its stages and its input and output bindings. It is additionally responsible for publishing the model online. The gathered information is later used when executing the model.

Fundamentally, the Python SDK allows MDs to declaratively define the desired inputs of their model with an object based component API. An MD will define a function for each step, and within each function, will define the components they wish to be rendered on a web-page. Each of these sequential functions will receive the input that an MC has provided via the interface. Additionally, the components also dictate the type of input that is expected from an MC. Knowing these types enables us to use the code defined by the SDK to create an API with input type definitions that can be programmatically executed.

Consider the VQG example listed above. Here, the MD will specify 3 functions. In the first function, the MD can use a \texttt{File.Upload} component to accept a specified number of image files (e.g. \texttt{File.Upload(max=5, accept=[".jpg", ".png"])}. This component will be rendered to a webpage. The MC then interacts with the component by uploading up to 5 image files. These files are now sent as input to the second function - accessible by the code the MD has written. In the second function, the MD will consume these images and run an object detector model on them. The object detector outputs a set of \texttt{detected\_objects} in the image. The MD can now specify an \texttt{Image.WithSelectMulti(images: List[np.array], detected\_objects: List[List[string]], max=2)} component to render a component which shows the N images alongside the n’th relevant sublist from \texttt{detected\_objects}. Once rendered, the MC can then select up to 2 detected objects per image. The selected objects are then sent to the 3rd step, where the MD will feed these objects to a language generator, finally returning the outputs with an \texttt{questions = model(images, selected\_objects); Image.View(images: List[array], questions: List[string])} component. 

A full example of the components we support thus-far are shown in Appendix \ref{app:api}. Now that we have motivated how users interact with the SDK, we discuss how we use the component specification to publish a model.

\begin{lstlisting}[language=json,firstnumber=1,caption={Example JSON metadata produced by the feather SDK. In the first step, we have a `File.Upload' component. The `props' attribute contains meta-information that the API and front-end will enforce. This information either comes from default values, or are specified by the MD as parameters for a component they use via the SDK. The `schema' attribute is the expected return type of that component. For the `Image.WithSelectMulti' component, we expect to recevie an object called `images'. Each image in images has a base64 string and filename assoicated to it, alongside an array of `attributes' which indicates the items that have been selected by the MC. \{...\} indicates information omitted for space reasons.},label={lst:metadata}]
{"steps": [
  {
    "name": "upload_images",
    "inputs": [
      {
        "component": "File.Upload",
        "props": {
          "max_files": 5,
          "min_files": 1,
          "title": "Upload images",
          "types": [
              ".jpg",
              ".png"
          ]
        },
        "schema": {...}
      }
    ]
  },
  {
    "name": "select_objects",
    "inputs": [
      {
        "component": "Image.WithSelectMulti",
        "props": {...},
        "schema": {
          "images": [
            {
              "attributes": [
                {
                  "item": "string",
                  "selected": "bool"
                }
              ],
                "data": "b64",
                "name": "string"
            }
          ]
        }
      }
    ]
  },
  {...}
]}
\end{lstlisting}

\subsection{Metadata}\label{sec:metadata}
The SDK generates JSON which contains the metadata of the MD’s model. The metadata includes, for each model state/step, the inputs and outputs types for each step, alongside component-specific configurations (e.g. number of files required for a `File.Upload' input). With the exception of the first step, the actual input and output data is dynamically communicated between the SDK and interface and thus not included in the metadata. Example metadata can be seen in Listing \ref{lst:metadata}.
The SDK captures model metadata about a model's inputs and outputs via Pythons's object introspection coupled with the strongly typed feather components defined in the SDK. The metadata captured is then used to generate the interactive UI for each input and output, as well as for data validation on the server side during model execution.

\subsection{Publishing}\label{sec:publishing}
Publishing is the process of uploading a model to the feather service, making it available in a user's library and for cloud based execution. An MD publishes their model using the feather SDK, which communicates with the feather service transparently. The following steps are executing during a publish:
\begin{enumerate}
    \item The SDK begins running the model locally at the first step. As the model executes, the metadata is generated for each step.
    \item The SDK retrieves a publishing token from the feather service. This step both validates the user's credentials and registers the model and model version if required.
    \item The SDK uploads the metadata JSON and the list of binary files to upload. The binary files are any files needed to run the model (eg. weights, data files, etc...). For each binary file the service returns a signed AWS S3 upload URL.
    \item The SDK uploads each binary file to the provided upload URL
    \item The SDK informs the feather service that upload is complete. The service validates each file against the metadata, and then transfers the model and binary files from the staging to the published storage.
\end{enumerate}

\subsection{Service}\label{sec:service}
The feather service runs as an AWS ECS container and provides support for user and model management, model publishing, as well as model execution. This small Go service exposes a REST API that is used by the Python SDK during publishing and by the feather web app during model execution.
The bulk of the service comprises validation code for both the publishing and the execution path. It is vital that all code is validated, all dependencies are tracked for the user uploaded python code, and all inputs are validated against the model metadata. Without this, the service cannot return precise error information to the MD, and multiple security vulnerabilities may be exploited.

Model versioning is also performed automatically by the service. As the MD publishes their model, the service automatically creates new versions (based on a `name' parameter the MD specifies during the publishing process) and preserves the previous version and all uploaded artifacts. This provides a rudimentary version control for MDs with simple roll-back and tagging. The MD can promote a specific version to the the live version visible by MCs while continuing to iterate on the model without fear of breaking anything. The versioning system is built on the fact that all resources published by MDs are immutable.

\subsection{Running the model}\label{sec:running}
Once a model has been published to the feather service, it can be executed via the feather web app. On the back-end, execution happens in an AWS Lambda in a security sandbox, as executing the model involves executing user provided Python code (which was uploaded during the publish phase).

In most cases, executing a multi-step model requires human interaction. Thus it cannot be achieved in one-shot. Hence we design the system such that each step in the model defines its inputs and outputs. Inputs are provided by the user, the model runs, and then outputs are returned. There is no requirement to execute steps in order or straight away; each step runs as a completely new AWS Lambda instance relying solely on its inputs \& compiled-in dependencies to function. In fact, this is exactly how feather supports API execution of models. Since the data transferred to and from the front-end calls an underlying API, we can simply expose this API to end users.

\begin{figure}[ht]
    \centering
    \includegraphics[width=\linewidth]{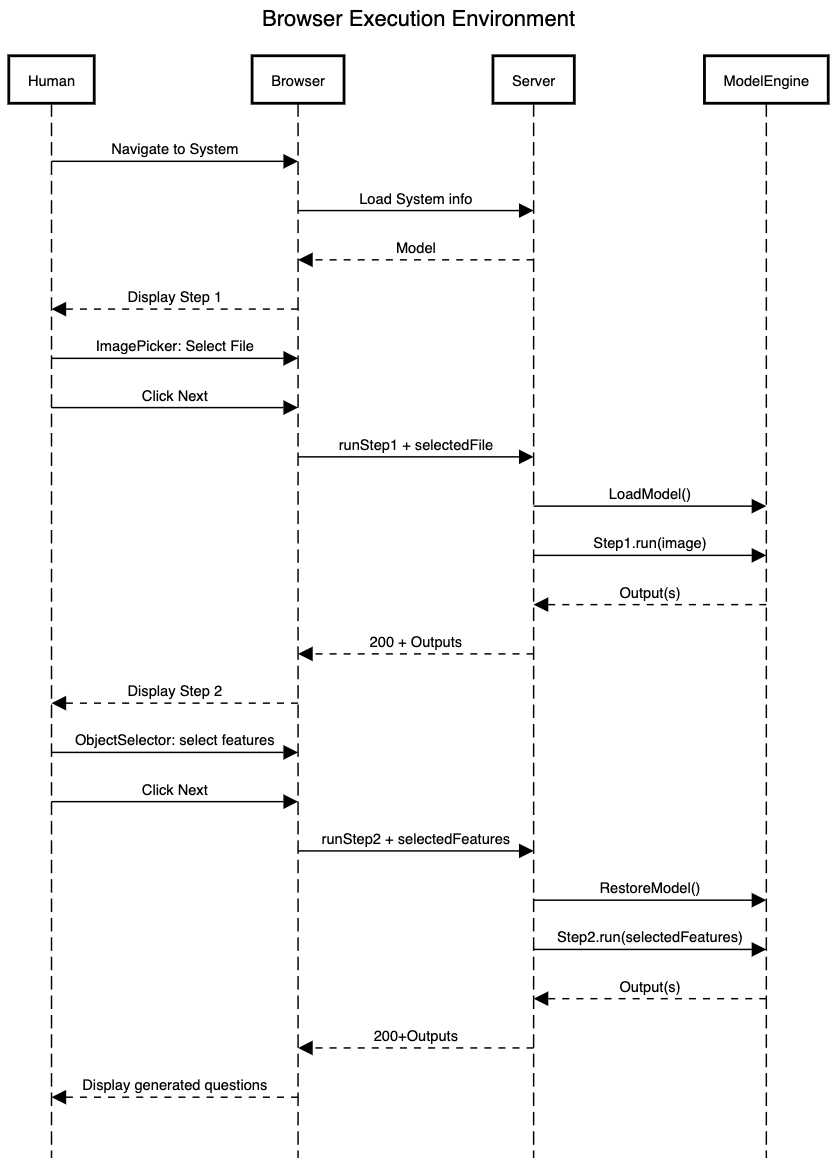}
    \caption{Sequence diagram for a multi-step model execution highlighting how user interaction is involved after each step.}
    \label{fig:sequence_diagram}
\end{figure}

 Figure \ref{fig:sequence_diagram} further details how our front-end interacts with users and the SDK. To be explicit, the front-end is where MCs will typically interact with the model (though they may also use the API). An example UI can be seen in Figure \ref{fig:script-and-ui}. To render this UI, we require a method which maps the components specified by the MD to web components. We achieve this by receiving the metadata from feather's back-end. As mentioned in Section \ref{sec:metadata}, the metadata contains an object which contains the components required for each step. Each component in each step additionally has metadata associated with it (e.g. `max: 5' for a File.Upload component). Our front-end loops over each step object in the metadata, and then loops over the components in each step, rendering each in the order specified. Once a user finishes interacting with a step, the data they enter is submitted and subsequently executed by the feather service.

\section{Areas for Improvement and Extensions}\label{sec:improvements}
In this section we discuss the shortcomings of our SDK as a whole. We provide details on how these shortcomings can be mitigated. Additionally, we outline extensions related to the SDK that could provide desired features. 

\subsection{Dependencies}
Through the Python code, an MD can import any package their model needs. The set of all directly and indirectly imported packages comprise the code dependencies of the model. After publishing, the MDs code is uploaded to feather's backend and needs to be able to run successfully in an AWS Lambda. As such the execution environment needs to contain the correct packages as well as versions. 
The current implementation of feather uses a baked Model Runner Lambda Image constructed manually to include a number of common Python packages. During model creation, the MD must only use the set of packages feather supports - although during our initial release, we often updated our image to include any new requested packages.

This is obviously not ideal and a better approach is to scan the packages used by a model during the publishing phase, subsequently uploading the dependency list and versions as part of the model metadata. The feather service could then maintain a registry of AWS Images, all created dynamically. Upon publishing a model, we could hash the package dependencies, and then look in the registry for a matching Execution Image. If one is found, then it can be used for the model any time it's executed. If one is not found, then once could be created dynamically and added to the registry.

The act of baking the dependencies into the Lambda image is to avoid downloading and installing anything during step execution. Avoiding this allows feather to execute a step in about 1 second.

\subsection{GPU Inference}
A worthwhile improvement is the support for GPUs during model execution/inference. The initial implementation of the feather model runner relied on executing Docker Containers on AWS Lambdas. Disregarding financial cost, there is technically nothing stopping a model runner using GPU Instances which would make inference much faster for large models. We envision that feather would support both CPU and GPU model runners, and allow MDs to specify at model publishing time via the SDK, what execution profile to use: low, medium or high performance; with high performance mapping to GPUs. This would then be factored into any future pricing model when executing the models.

\subsection{Richer Model Definitions}
Currently, models which are defined by our SDK can only be executed in a linear fashion for the number of steps that have been defined. This prevents programs that may want to conditionally execute a model-step, or those that run a model-step recursively. One example may be a multi-step image in-painting model where an MC in-paints a background before moving on to in-paint the foreground. Here, the user may want to iteratively regenerate the background. Once they are satisfied, the conditional branch the MD specified will allow them start generating the foreground. 

Enabling this may be achieved in multiple ways. We could provide a flowchart-like visual interface for MDs to create the execution logic of each step. Alternatively we can we could introduce new types of model steps: a conditional step which switches between different functions based on the input, and a recursive step which either allows the user to specify either the number of iterations, or a break condition.

\subsection{Fine-tuning as a Service}
A further extension to the SDK can enable automatic fine-tuning of a base model. In this example, the MD can specify the code to continue training their model alongside a schema of accepted data. feather's UI/API endpoint can then accept training data from the MC in the format of the defined schema. The model runner can then generate a new fine-tuned model by training the base model against the provided dataset.

\section{Conclusion}
In this paper we introduced feather: an SDK that allows model developers to declaratively share and deploy their models. This was motivated by the lack of reproducibility in machine learning, and the pragmatic difficulties of a model developer exposing their model for easy inference. We outlined the implementation details of our SDK, and how the deployed models may be executed either via a UI or an API. Finally, we discussed additional improvements that could be made to the SDK.

\bibliography{references}
\bibliographystyle{icml2022}

\newpage
\appendix
\section{feather's component API}\label{app:api}
    
The SDK exposes a Object Based Component API. Components allow MCs to declaratively define what types of data they are working with, along with easy semantics for controlling the UI behaviour for each component. Each Component in the SDK maps to UI controls, which allows models created through the feather SDK to be interacted with both programatically (via Python) or visually via the Web Portal. 

Below are the components that are currently supported by feather:
\begin{lstlisting}
class File
  def Upload()
  def Download()
  
class Text
  def In()
  def View() 
  
class List
  def SelectOne()
  def SelectMulti()
  
class Image
  def WithSelectOne()
  def WithSelectMulti()
  def WithTextIn()
  def View()
  
class Document
  def WithTextIn()
  def View()
\end{lstlisting}

\end{document}